\documentclass[11pt]{article}
\usepackage{colacl}

\title{Evaluating a Focus-Based Approach to Anaphora Resolution}
\author{Saliha Azzam, Kevin Humphreys \and Robert Gaizauskas\\
\texttt{\{s.azzam,k.humphreys,r.gaizauskas\}@dcs.shef.ac.uk}\\
Department of Computer Science, University of Sheffield\\
Regent Court, Portobello Road\\
Sheffield\ \ S1 4DP\ \ UK}

\begin{document}

\maketitle

\bibliographystyle{acl} 
\begin{abstract}
We present an approach to anaphora resolution based on a focusing
  algorithm, and implemented within an existing MUC (Message
  Understanding Conference) Information Extraction system, allowing
  quantitative evaluation against a substantial corpus of annotated
  real-world texts. Extensions to the basic focusing mechanism can be
  easily tested, resulting in refinements to the mechanism and
  resolution rules. Results show that the focusing algorithm is highly
  sensitive to the quality of syntactic-semantic analyses, when
  compared to a simpler heuristic-based approach.
\end{abstract}

\section{Introduction}
\label{sec:intro}

Anaphora resolution is still present as a significant linguistic
problem, both theoretically and practically, and interest has recently
been renewed with the introduction of a quantitative evaluation
regime as part of the Message Understanding Conference (MUC)
evaluations of Information Extraction (IE) systems \cite{Gri96}.  This
has made it possible to evaluate different (implementable) theoretical
approaches against sizable corpora of real-world texts, rather than the
small collections of artificial examples typically discussed in the
literature.

This paper\footnote{This work was carried out in the context of the EU
  AVENTINUS project (Thurmair, 1996), which aims to develop a
  multilingual IE system for drug enforcement, and including a
  language-independent coreference mechanism (Azzam et al., 1998).}
describes an evaluation of a focus-based approach to pronoun
resolution (not anaphora in general), based on an extension of
Sidner's algorithm \cite{Sid81} proposed in \cite{Azz96}, with further
refinements from development on real-world texts.  The approach is
implemented within the general coreference mechanism provided by the
LaSIE (Large Scale Information Extraction) system \cite{Gai95b} and
\cite{Hum98}, Sheffield University's entry in the MUC-6 and 7
evaluations.

\section{Focus in Anaphora Resolution}
\label{sec:focus}

The term \emph{focus}, along with its many relations such as
\emph{theme}, \emph{topic}, \emph{center}, etc., reflects an intuitive
notion that utterances in discourse are usually `about' something.
This notion has been put to use in accounts of numerous linguistic
phenomena, but it has rarely been given a firm enough definition to
allow its use to be evaluated.  For anaphora resolution, however,
stemming from Sidner's work, focus has been given an algorithmic
definition and a set of rules for its application.  Sidner's approach
is based on the claim that anaphora generally refer to the current
discourse focus, and so modelling changes in focus through a discourse
will allow the identification of antecedents.

The algorithm makes use of several \emph{focus registers} to represent
the current state of a discourse: \emph{CF}, the current focus;
\emph{AFL}, the alternate focus list, containing other candidate foci;
and \emph{FS}, the focus stack.  A parallel structure to the
\emph{CF}, \emph{AF} the actor focus, is also set to deal with 
agentive pronouns.  The algorithm updates these registers after each
sentence, confirming or rejecting the current focus.  A set of
\emph{Interpretation Rules} (\emph{IR}s) applies whenever an anaphor
is encountered, proposing potential antecedents from the registers,
from which one is chosen using other criteria: syntactic, semantic,
inferential, etc.

\subsection{Evaluating~Focus-Based~Approaches}
  
Sidner's algorithmic account, although not exhaustively specified, has
lead to the implementation of focus-based approaches to anaphora
resolution in several systems, e.g.\ PIE \cite{Lin95}.  However,
evaluation of the approach has mainly consisted of manual analyses of
small sets of problematic cases mentioned in the literature.  Precise
evaluation over sizable corpora of real-world texts has only recently
become possible, through the resources provided as part of the MUC
evaluations.

\section{Coreference in LaSIE}
\label{sec:lasie-coref}

The LaSIE system \cite{Gai95b} and \cite{Hum98}, has been designed as
a general purpose IE system which can conform to the MUC task
specifications for named entity identification, coreference
resolution, IE template element and relation identification, and the
construction of scenario-specific IE templates.  The system is
basically a pipeline architecture consisting of tokenisation, sentence
splitting, part-of-speech tagging, morphological stemming, list
lookup, parsing with semantic interpretation, proper name matching,
and discourse interpretation.  The latter stage constructs a discourse
model, based on a predefined domain model, using the, often partial,
semantic analyses supplied by the parser.

The domain model represents a hierarchy of domain-relevant concept
nodes, together with associated properties.  It is expressed in the XI
formalism \cite{Gai95} which provides a basic inheritance mechanism
for property values and the ability to represent multiple
classificatory dimensions in the hierarchy.  Instances of concepts
mentioned in a text are added to the domain model, populating it to
become a text-, or discourse-, specific model.

Coreference resolution is carried out by attempting to merge each
newly added instance, including pronouns, with instances already
present in the model.  The basic mechanism is to examine, for each
new-old pair of instances: semantic type consistency/similarity in the
concept hierarchy; attribute value consistency/similarity, and a set
of heuristic rules, some specific to pronouns, which can act to rule
out a proposed merge.  These rules can refer to various lexical,
syntactic, semantic, and positional information about instances. The
integration of the focus-based approach replaces the heuristic rules
for pronouns, and represents the use of LaSIE as an evaluation
platform for more theoretically motivated algorithms.  It is possible
to extend the approach to include definite NPs but, at present, the
existing rules are retained for non-pronominal anaphora in the MUC
coreference task: proper names, definite noun phrases and bare nouns.

\section{Implementing Focus-Based Pronoun Resolution in LaSIE}
\label{sec:lasie-focus}

Our implementation makes use of the algorithm proposed in
\cite{Azz96}, where \emph{elementary events} (\emph{EE}s, effectively
simple clauses) are used as basic processing units, rather than
sentences.  Updating the focus registers and the application of
interpretation rules (\emph{IR}s) for pronoun resolution then takes
place after each \emph{EE}, permitting intrasentential
references.\footnote{An important limitation of Sidner's algorithm,
  noted in \cite{Azz96}, is that the focus registers are only updated
  after each sentence.  Thus antecedents proposed for an anaphor in
  the current sentence will always be from the previous sentence or
  before and intrasentential references are impossible.}  In
addition, an initial `expected focus' is determined based on the first
\emph{EE} in a text, providing a potential antecedent for any pronoun
within the first \emph{EE}.

Development of the algorithm using real-world texts resulted in
various further refinements to the algorithm, in both the \emph{IR}s
and the rules for updating the focus registers.  The following
sections describe the two rules sets separately, though they are
highly interrelated in both development and processing.

\subsection{Updating the Focus}
\label{sec:updating}

The algorithm includes two new focus registers, in addition to those
mentioned in section~\ref{sec:focus}: \emph{AFS}, the actor focus
stack, used to record previous \emph{AF} (actor focus) values and so
allow a separate set of \emph{IR}s for \emph{agent} pronouns (animate
verb subjects); and \emph{Intra-AFL}, the intrasentential alternate
focus list, used to record candidate foci from the current
\emph{EE} only.

In the space available here, the algorithm is best described through
an example showing the use of the registers.  This example is taken
from a New York Times article in the MUC-7 training corpus on aircraft
crashes:\\[-1ex]

{\it State Police said witnesses told them the propeller was not
turning as the plane descended quickly toward the highway in
Wareham near Exit 2. It hit a tree.}\\[-1ex]

\noindent \textbf{EE-1}: \textit{State Police said} \texttt{tell\_event}

An `expected focus' algorithm applies to initialise the
registers as follows:\\
\emph{CF} (current focus) = \texttt{tell\_event}\\
\emph{AF} (actor focus) = \textit{State Police}\\
\emph{Intra-AFL} remains empty because EE-1 contains no other
candidate foci.  No other registers are affected by the expected
focus.  No pronouns occur in EE-1 and so no \emph{IR}s apply.\\[-1ex]

\noindent \textbf{EE-2}: \textit{witnesses told them}

The \emph{Intra-AFL} is first initialised with all (non-pronominal)
candidate foci in the EE:\\
\emph{Intra-AFL} = \textit{witnesses}\\
The \emph{IR}s are then applied to the first pronoun, \textit{them},
and, in this case, propose the current \emph{AF}, \textit{State
  Police}, as the antecedent.  The \emph{Intra-AFL} is immediately
updated to add the antecedent:\\
\emph{Intra-AFL} = \textit{State Police}, \textit{witnesses}\\
EE-2 has a pronoun in `thematic' position, `theme' being either the
object of a transitive verb, or the subject of an intransitive or the
copula (following \cite{Gru76}).  Its antecedent therefore becomes the
new \emph{CF}, with the previous value moving to the \emph{FS}.  EE-2
has an `agent', where this is an animate verb subject (again as in
\cite{Gru76}), and this becomes the new \emph{AF}.  Because the old
\emph{AF} is now the \emph{CF}, it is not added to the \emph{AFS} as
it would be otherwise.  After each EE the \emph{Intra-AFL} is added to
the current \emph{AFL}, excluding the \emph{CF}.  The state after EE-2 is then:\\
\emph{CF} = \textit{State Police}\ \ \ \ \emph{AF} = \textit{witnesses}\\
\emph{FS} = \texttt{tell\_event}\ \ \ \ \ \emph{AFL} = \textit{witnesses}\\[-1ex]

\noindent \textbf{EE-3}: \textit{the propeller was not turning}

The \emph{Intra-AFL} is reinitialised with candidate foci from this
EE:\\
\emph{Intra-AFL} = \textit{propeller}\\
No pronouns occur in EE-3 and so no \emph{IR}s apply.  The `theme',
\textit{propeller} here because of the copula, becomes the new
\emph{CF} and the old one is added to the \emph{FS}.  The \emph{AF}
remains unchanged as the
current \emph{EE} lacks an agent:\\
\emph{CF} = \textit{propeller}\\
\emph{AF} = \textit{witnesses}\\
\emph{FS} = \textit{State Police}, \texttt{tell\_event}\\
\emph{AFL} = \textit{propeller}, \textit{witnesses}\\[-1ex]

\noindent \textbf{EE-4}: \textit{the plane descended}

\noindent \emph{Intra-AFL} = \textit{the plane}\\
\emph{CF} = \textit{the plane} (theme)\\
\emph{AF} = \textit{witnesses} (unchanged)\\
\emph{FS} = \textit{propeller}, \textit{State Police}, \texttt{tell\_event}\\
\emph{AFL} = \textit{the plane}, \textit{propeller}, \textit{witnesses}\\
In the current algorithm the \emph{AFL} is reset at this point,
because EE-4 ends the sentence.\\[-1ex]

\noindent \textbf{EE-5}: \textit{it hit a tree}

\noindent \emph{Intra-AFL} = \textit{a tree}\\
The \emph{IR}s resolve the pronoun \textit{it} with the \emph{CF}:\\
\emph{CF} = \textit{the plane} (unchanged)\\
\emph{AF} = \textit{witnesses} (unchanged)\\
\emph{FS} = \textit{propeller}, \textit{State Police}, \texttt{tell\_event}\\
\emph{AFL} = \textit{a tree}

\subsection{Interpretation Rules}
\label{sec:rules}

Pronouns are divided into three classes, each with a distinct set of
\emph{IR}s proposing antecedents:
\paragraph{Personal pronouns acting as agents (animate subjects):} 
  (e.g.\ \textit{he} in {\it Shotz said he knew the pilots})
  \emph{AF} proposed initially, then animate members of \emph{AFL}.
\paragraph{Non-agent pronouns:} 
  (e.g.\ \textit{them} in EE-2 above and \textit{it} in EE-5)
  \emph{CF} proposed initially, then members of the \emph{AFL} and
  \emph{FS}. 
\paragraph{Possessive, reciprocal and reflexive pronouns (\emph{PRR}s):} 
  (e.g.\ \textit{their} in {\it the brothers had left and were on
    their way home})
  Antecedents proposed from the \emph{Intra-AFL}, allowing intra-EE
    references.  

Antecedents proposed by the \emph{IR}s are accepted or rejected based
on their semantic type and feature compatibility, using the semantic
and attribute value similarity scores of LaSIE's existing coreference
mechanism.

\section{Evaluation with the MUC Corpora}
\label{sec:evaluation}

As part of MUC \cite{Gri96}, coreference resolution was evaluated as a
sub-task of information extraction, which involved negotiating a
definition of coreference relations that could be reliably evaluated.
The final definition included only `identity' relations between text
strings: proper nouns, common nouns and pronouns.  Other possible
coreference relations, such as `part-whole', and non-text strings
(zero anaphora) were excluded.

The definition was used to manually annotate several corpora of
newswire texts, using SGML markup to indicate relations between text
strings.  Automatically annotated texts, produced by systems using the
same markup scheme, were then compared with the manually annotated
versions, using scoring software made available to MUC participants,
based on \cite{Vil95}.

The scoring software calculates the standard Information Retrieval
metrics of `recall' and `precision',\footnote{Recall is a measure of
  how many correct (i.e.\ manually annotated) coreferences a system
  found, and precision is a measure of how many coreferences that the
  system proposed were actually correct.  For example, with 100
  manually annotated coreference relations in a corpus and a system
  that proposes 75, of which 50 are correct, recall is then $50/100$
  or $50\%$ and precision is $50/75$ or $66.7\%$.}  together with an
overall \textit{f}-measure.  The following section presents the
results obtained using the corpora and scorer provided for MUC-7
training (60 texts, average 581 words per text, 19 words per sentence)
and evaluation (20 texts, average 605 words per text, 20 words per
sentence), the latter provided for the formal MUC-7 run and kept blind
during development.

\section{Results}
\label{sec:results}

The MUC scorer does not distinguish between different classes of
anaphora (pronouns, definite noun phrases, bare nouns, and proper
nouns), but baseline figures can be established by running the LaSIE
system with no attempt made to resolve any pronouns:
\begin{verbatim}
    Corpus    Recall  Precision    f
  Training:    42.4%    73.6%    52.6%
  Evaluation:  44.7%    73.9%    55.7%
\end{verbatim}

LaSIE with the simple pronoun resolution heuristics of the
non-focus-based mechanism achieves the following:
\begin{verbatim}
    Corpus    Recall  Precision    f
  Training:    58.2%    71.3%    64.1%
  Evaluation:  56.0%    70.2%    62.3%
\end{verbatim}
 
showing that more than three quarters of the estimated 20\% of pronoun
coreferences in the corpora are correctly resolved with only a minor
loss of precision. 

LaSIE with the focus-based algorithm achieves the following:
\begin{verbatim}
    Corpus    Recall  Precision    f
  Training:    55.4%    70.3%    61.9%
  Evaluation:  53.3%    69.7%    60.4%
\end{verbatim}

which, while demonstrating that the focus-based algorithm is
applicable to real-world text, does question whether the more complex
algorithm has any real advantage over LaSIE's original simple
approach.

The lower performance of the focus-based algorithm is mainly due to an
increased reliance on the accuracy and completeness of the grammatical
structure identified by the parser.  For example, the resolution of a
pronoun will be skipped altogether if its role as a verb argument is
missed by the parser.  Partial parses will also affect the
identification of \textit{EE} boundaries, on which the focus update
rules depend.  For example, if the parser fails to attach a
prepositional phrase containing an antecedent, it will then be missed
from the focus registers and so the \textit{IR}s (see \cite{Azz95b}).
The simple LaSIE approach, however, will be unaffected in this case.

Recall is also lost due to the more restricted proposal of candidate
antecedents in the focus-based approach.  The simple LaSIE approach
proposes antecedents from each preceding paragraph until one is
accepted, while the focus-based approach suggests a single fixed set.

From a theoretical point of view, many interesting issues appear with
a large set of examples, discussed here only briefly because of lack
of space.  Firstly, the fundamental assumption of the focus-based
approach, that the focus is favoured as an antecedent, does not always
apply.  For example:\\[-1ex]

{\it In June, a few weeks before the crash of TWA Flight 800, leaders
  of several Middle Eastern terrorist organizations met in Teheran to
  plan terrorist acts.  Among \underline{them} was the PFL of
  Palestine, an organization that has been linked to airplane bombings
  in the past.}\\[-1ex]

Here, the pronoun {\it them} corefers with {\it organizations} rather than
the focus {\it leaders}.  Additional information will be required to
override the fundamental assumption.

Another significant question is when sentence focus changes.  In
our algorithm, focus changes when there is no reference (pronominal or
otherwise) to the current focus in the current {\it EE}.  In the
example used in section~4.1, this causes the focus at the end of the
first sentence to be that of the last \textit{EE} in that sentence,
thus allowing the pronoun {\it it} in the subsequent sentence to be
correctly resolved with {\it the plane}.  However in the example
below, the focus of the first \textit{EE} ({\it the writ}) is the
antecedent of the pronoun \textit{it} in the subsequent sentence,
rather than the focus from the last \textit{EE} (\textit{the \ldots
  flight}):\\[-1ex]

{\it The writ is for ``damages'' of seven passengers who died when the
  Airbus A310 flight crashed. It claims the deaths were caused by
  negligence.} \\[-1ex]
  
Updating focus after the complete sentence, rather than each
\textit{EE}, would propose the correct antecedent in this case.
However neither strategy has a significant overall advantage in
our evaluations on the MUC corpora.

Another important factor is the priorities of the Interpretation
Rules. For example, when a personal pronoun can corefer with both {\it
  CF} and {\it AF}, \textit{IR}s select the {\it CF} first in our
algorithm.  However, this priority is not fixed, being based only on
the corpora used so far, which raises the possibility of automatically
acquiring \textit{IR} priorities through training on other corpora.

\section{Conclusion}
\label{sec:conc}

A focus-based approach to pronoun resolution has been implemented
within the LaSIE IE system and evaluated on real-world texts.  The
results show no significant preformance increase over a simpler
heuristic-based approach.  The main limitation of the focus-based
approach is its reliance on a robust syntactic/semantic analysis to
find the focus on which all the \emph{IR}s depend.  Examining
performance on the real-world data also raises questions about the
theoretical assumptions of focus-based approaches, in particular
whether focus is always a favoured antecedent, or whether this
depends, to some extent, on discourse style.

Analysing the differences in the results of the focus- and
non-focus-based approaches, does show that the focus-based rules are
commonly required when the simple syntactic and semantic rules propose
a set of equivalent antecedents and can only select, say, the closest
arbitrarily.  A combined approach is therefore suggested, but whether
this would be more effective than further refining the resolution
rules of the focus-based approach, or improving parse results and
adding more detailed semantic constraints, remains an open question.



\end{document}